%% file: main.tex
\documentclass[10pt,twocolumn,letterpaper]{article}

\usepackage[pagenumbers]{cvpr} % To force page numbers, e.g. for an arXiv version

\input{preamble}
\definecolor{cvprblue}{rgb}{0.21,0.49,0.74}
\definecolor{pink}{rgb}{0.929,0.008,0.549}
\usepackage[
    pagebackref,
    breaklinks,
    colorlinks=true,
    linkcolor=cvprblue,
    citecolor=cvprblue,
    filecolor=cvprblue,
    urlcolor=pink
]{hyperref}

\usepackage{enumitem}
\usepackage{xcolor}
\usepackage{url}
\usepackage{caption}
\definecolor{cvprblue}{rgb}{0.21,0.49,0.74}
\definecolor{brightblue}{rgb}{0.0196, 0.4431, 0.7412}
\definecolor{icmlblue}{rgb}{0,0.08,0.45}
\definecolor{darkred}{rgb}{0.75,0,0}
\definecolor{purple}{rgb}{0.486, 0.153, 0.902}
\definecolor{forestgreen}{rgb}{0, 0.502, 0.502}
\definecolor{grey}{rgb}{0.45, 0.45, 0.45}

\usepackage[normalem]{ulem}
\usepackage{url}                                 
\usepackage{graphicx}    
\usepackage{makecell}
\newcommand{\bluetext}[1]{\textcolor{blue}{#1}} 
\usepackage{booktabs} 
\usepackage{caption} % for captions
\usepackage{float} % for positioning  
\usepackage{tabularray}
\usepackage{multirow}
\usepackage{amsmath}
\usepackage[ruled,vlined,linesnumbered]{algorithm2e}

\usepackage{amssymb}
\usepackage{fontawesome5}
\usepackage[ruled,vlined]{algorithm2e} 
\usepackage{lineno}
\usepackage{pifont}
\usepackage{dsfont}
\usepackage{tikz}

\usepackage{caption}
\usepackage{subcaption}

\def\paperID{*****} % *** Enter the Paper ID here
\def\confName{CVPR}
\def\confYear{2026}

\title{No Adaptation Without Observation:
Observability-Constrained \\Test-Time Prompt Tuning for LiDAR Semantic Segmentation}

\author{
Linlian Jiang$^{1,2}$ \quad
Wentao Ju$^{1}$ \quad
Sadman Rakib Pinon$^{1,2}$ \quad
Jianwei Xian$^{1}$ \quad
Zhixiang Chi$^{3}$ \\
Xinxin Zuo$^{1}$\thanks{Corresponding authors.} \quad Yang Wang$^{1,2}$\footnotemark[1] \\ \\
$^{1}$Concordia University \qquad
$^{2}$Mila -- Quebec AI Institute \qquad
$^{3}$University of Toronto \\
{\small\tt \{linlian.jiang,wentao.ju,sadmanrakib.pinon,jianwei.xian\}@mail.concordia.ca} \\
{\small\tt \{zhxchi\}@ece.utoronto.ca, \{xinxin.zuo,yang.wang\}@concordia.ca}
}

\begin{document}
\maketitle
\begin{tikzpicture}[remember picture,overlay]

\node[
    anchor=south east,
    xshift=-5.5em,
    yshift=3.7em
] at (current page.south east) {

\begin{minipage}{0.85\linewidth}
\raggedleft\footnotesize\itshape
``A wise man, therefore, proportions his belief to the evidence.''\\[0.7ex]
--- David Hume
\end{minipage}

};

\end{tikzpicture}

\input{sec/0_abstract}    
\input{sec/1_intro}

\input{sec/2_related}

\input{sec/3_method}
\input{sec/4_exps}
\input{sec/con}
{
    \small
    \bibliographystyle{ieeenat_fullname}
    \bibliography{main}
}


\end{document}

%% file: sec/0_abstract.tex
\begin{abstract}
LiDAR semantic segmentation often degrades under real-world deployment
due to evolving sensing conditions, while collecting new annotations
for retraining is impractical.
Test-time adaptation (TTA) updates model parameters online using pseudo-label supervision,
but directly applying standard TTA strategies to LiDAR data is challenging.
Because pseudo-label reliability is spatially heteroscedastic under
range-dependent sparsity and occlusion,
uniform updates on globally shared parameters
can inject unstable gradients and destabilize adaptation.
We propose a geometry-constrained test-time prompt tuning framework
for LiDAR semantic segmentation.
Our method estimates per-location sensing reliability from
depth-consistent beam terminations and neighborhood support,
and uses it to reweight spatial supervision.
Adaptation is confined to lightweight prompt adapters inserted
into a frozen backbone, with spatial gating to prevent unreliable
regions from perturbing globally shared representations.
A temporally smoothed prototype alignment strategy further stabilizes
online updates by accumulating reliable semantic evidence over time.
Experiments on standard LiDAR benchmarks demonstrate
improved adaptation stability and segmentation performance
under deployment variations without additional annotations. 
Code is available at \href{https://linlianjiang.github.io/noob/}{our project page}.
\end{abstract}

%% file: sec/1_intro.tex
\section{Introduction}

\begin{figure}[t]
    \centering
    \includegraphics[width=1\linewidth]{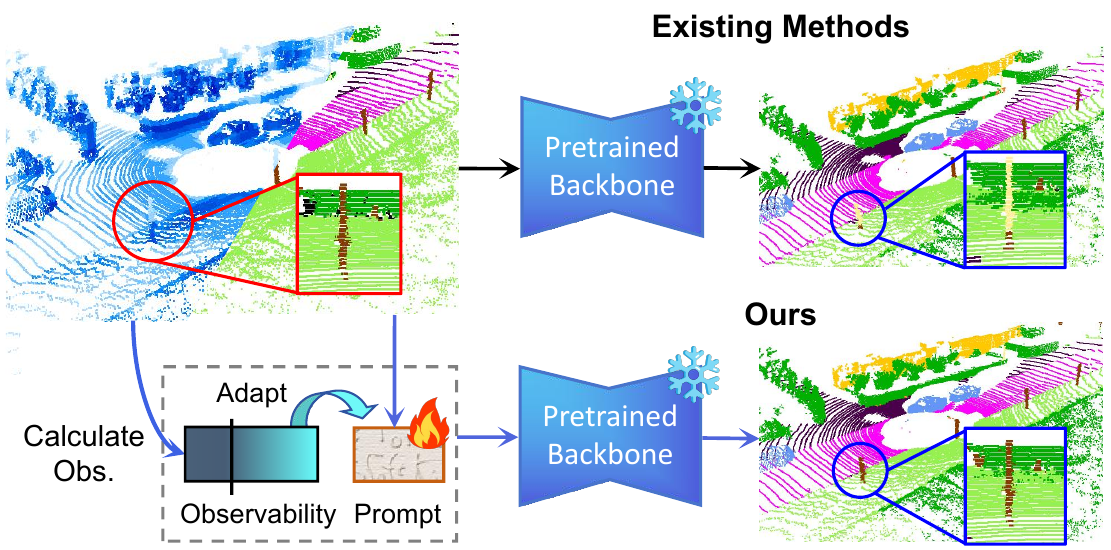}
\caption{
We compute a geometry-aware observability score that quantifies sensing reliability in LiDAR scenes
and use it to modulate prompt-based adaptation within a frozen backbone,
so that parameter updates are driven only by geometrically supported evidence.
}
    \label{fig:teaser}
\end{figure}

LiDAR semantic segmentation assigns semantic labels to 3D points
and serves as a core module for downstream
perception~\cite{ye2023lidarmultinet}, prediction~\cite{peri2022forecasting}, and planning~\cite{hu2023planning}
in autonomous driving and robotic systems.
Driven by large-scale annotated datasets~\cite{behley2019semantickitti,caesar2020nuscenes},
modern deep networks, including sparse voxel-based 3D models~\cite{zhu2021cylindrical,lai2023spherical}
and efficient range-view projection-based architectures~\cite{milioto2019rangenet++,zhang2020polarnet,ando2023rangevit,zheng2024spherical,xu2025frnet},
achieve strong performance under supervised training.
However, maintaining this performance during real-world deployment
remains challenging as sensing conditions vary across scenes and frames,
and collecting new dense annotations for retraining is impractical.

Test-time adaptation (TTA)~\cite{wang2020tent,wang2022continual}
offers a practical mechanism to cope with deployment variations
by updating model parameters online using pseudo-label supervision.
However, directly applying standard TTA strategies to LiDAR data
remains non-trivial.
Most existing TTA methods update shared parameters
using predictions from the entire scene,
implicitly assuming that pseudo-labels across spatial locations
are equally reliable.
While this assumption is often acceptable in dense image domains~\cite{wang2020tent},
where observations are spatially continuous,
it becomes questionable in LiDAR sensing,
where geometric observability varies across space.
Geometric evidence varies substantially
due to range-dependent sparsity, occlusion, and ambiguous returns~\cite{milioto2019rangenet++,zhu2021cylindrical}.
As a result, supervision is inherently heteroscedastic,
and pseudo-label reliability becomes spatially uneven,
creating optimization signals with highly variable reliability.

This spatial heterogeneity becomes critical under parameter sharing.
During adaptation, gradients from all spatial locations
are accumulated into globally shared weights.
Consequently, unreliable supervision from a subset of regions
can bias representations across the entire scene~\cite{niu2022efficient}.
One common mitigation strategy is to restrict adaptation
to a lightweight parameter subspace,
as in prompt-based parameterization~\cite{jia2022visual,yang2022prompt},
where the backbone is frozen to protect pre-trained representations
and reduce adaptation degrees of freedom.
Yet, the learned prompts remain globally shared and spatially uniform,
allowing unreliable supervision to couple across regions
through these shared parameters.
This residual global sensitivity highlights the need
for spatially aware modulation
to stabilize LiDAR test-time adaptation.

To address the instability caused by spatially heterogeneous pseudo-label supervision,
we propose a geometry-constrained test-time prompt tuning framework
for LiDAR semantic segmentation.
The framework is guided by \textbf{a single geometric principle}:
\emph{no adaptation without observation} (Fig.~\ref{fig:teaser} illustrates this idea),
ensuring that parameter updates are driven only by geometrically supported evidence.
Specifically, this principle is enforced through three complementary components.
First, we introduce a geometry-aware observability mechanism
that estimates sensing reliability at each spatial location
using depth-consistent beam terminations and multi-scale neighborhood support.
This score serves as a continuous validity signal
that quantifies local geometric support.
Second, we confine adaptation to lightweight prompt adapters
within a frozen backbone for stable test-time updates.
Prompt-induced residual updates are modulated by the observability score,
so that only geometrically supported regions contribute
to shared parameter updates.
Third, we incorporate a temporally smoothed prototype alignment mechanism
that accumulates observability-filtered semantic evidence over time.
The resulting prototype-aligned targets regularize prompt updates
and suppress prediction flicker under intermittent geometric support.
Together, these components enforce geometry-aware adaptation
both spatially and temporally,
enabling stable and efficient LiDAR test-time adaptation
under evolving sensing conditions.

Our contributions are summarized as follows:
\begin{itemize}

\item We propose a geometry-aware observability operator that quantifies sensing reliability from beam terminations and neighborhood support, enabling spatially grounded supervision for LiDAR test-time adaptation.

\item We develop a geometry-constrained prompt tuning framework in which observability explicitly modulates prompt-induced updates within a frozen backbone, mitigating unreliable supervision under shared parameters.

\item We introduce an observability-filtered temporal prototype alignment mechanism that accumulates reliable semantic evidence over time and stabilizes online adaptation by suppressing prediction flicker.

\end{itemize}

%% file: sec/2_related.tex
\section{Related Work}

\noindent\textbf{LiDAR Semantic Segmentation.}
Existing LiDAR segmentation methods are generally categorized into
voxel-based 3D approaches and projection-based 2D approaches.
Voxel-based methods, including SPVCNN~\cite{tang2020searching},
Cylinder3D~\cite{zhu2021cylindrical},
and SphereFormer~\cite{lai2023spherical},
operate directly in 3D space and model sparse volumetric structures
with strong geometric locality, often achieving high accuracy
at increased computational cost.
Projection-based methods such as RangeNet++~\cite{milioto2019rangenet++},
PolarNet~\cite{zhang2020polarnet}, RangeViT~\cite{ando2023rangevit},
and SFCNet~\cite{zheng2024spherical}
transform point clouds into structured range-view representations,
enabling efficient dense convolution and fast inference.
FRNet~\cite{xu2025frnet} further explores frustum-based representations
to balance geometric fidelity and scalability.
Despite their strong supervised performance,
these models are typically trained offline and
lack mechanisms for online adaptation under deployment shifts.

\noindent\textbf{Visual Prompt Tuning.}
Prompt learning was originally proposed in natural language processing
as a parameter-efficient strategy to adapt large pre-trained models
while keeping most parameters frozen.
As part of the broader family of parameter-efficient fine-tuning (PEFT) methods~\cite{liu2023pre,liu2021p},
prompt-based adaptation has been extended to vision-language models
such as CLIP~\cite{radford2021learning},
and further developed into continuous and conditional prompting schemes~\cite{zhou2022learning,zhou2022conditional}.
In computer vision, prompt tuning has emerged as an effective alternative
to full fine-tuning for transformer-based models~\cite{jia2022visual,park2024fair},
and has recently been explored for point cloud understanding
and multi-modal learning~\cite{zhou2024dynamic,chi2025learning}.
However, existing prompt-based approaches are primarily designed
for supervised or offline adaptation.
Crucially, their learned prompts remain globally shared
and spatially uniform across spatial locations.
Such formulations do not account for the spatially heterogeneous reliability
inherent to LiDAR sensing,
and therefore lack explicit geometry-constrained mechanisms
for stable test-time adaptation.

\begin{figure*}[t] \centering \includegraphics[width=1\linewidth]{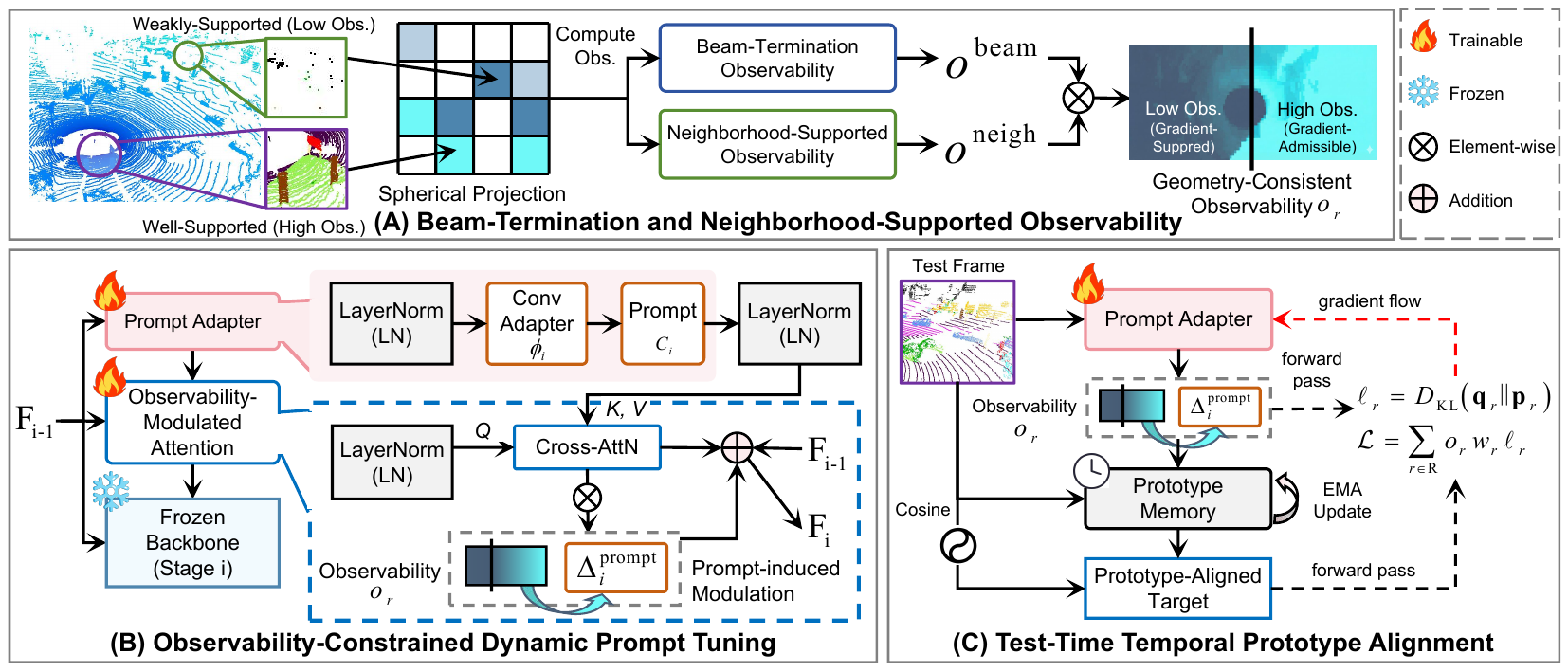} \caption{Geometry-constrained test-time adaptation.
Observability $o_r$ is computed from beam-termination and neighborhood support (A),
modulates prompt-induced residual updates within a frozen backbone (B),
and regularizes adaptation through temporally smoothed prototype alignment (C).} \label{fig:structure} \end{figure*}

\noindent\textbf{Test-Time Adaptation.}
Test-time adaptation (TTA) mitigates distribution shift
by updating model parameters at inference without labeled targets,
allowing pre-trained models to adapt to and personalize
the current test environment through self-generated supervision.
Early TTA methods optimize entropy minimization
or adapt normalization statistics at test time~\cite{wang2020tent,wang2022continual}.
Subsequent works explore continual and online adaptation strategies
to handle non-stationary target distributions
and mitigate error accumulation~\cite{niu2022efficient}.

In 3D LiDAR segmentation, recent approaches such as
GIPSO~\cite{saltori2022gipso} and HGL~\cite{zou2024hgl}
leverage geometric cues for online adaptation.
However, most existing methods implicitly treat supervision
from all spatial locations as equally reliable
and allow all regions to contribute uniformly
to parameter updates.
This assumption is problematic in sparse 3D sensing,
where geometric support is inherently uneven
due to range sparsity and occlusion.
Updates driven by weakly supported regions
can introduce unstable gradients
that corrupt globally shared representations,
leading to severe parameter drift.

%% file: sec/3_method.tex
\section{Method}

\noindent\textbf{Task Formulation.}
LiDAR semantic segmentation assigns a semantic label to each point in a point cloud.
Given a point cloud $P=\{p_i\}_{i=1}^N$, where $p_i \in \mathbb{R}^3$,
the goal is to learn a mapping
$f_\theta : \{p_i\}_{i=1}^N \rightarrow \{y_i\}_{i=1}^N,\; y_i \in K,$
where $K=\{1,\dots,C\}$ denotes the semantic categories.

\noindent\textbf{Network Structure and Adaptation Principle.}
We adopt FRNet~\cite{xu2025frnet} as the segmentation backbone.
Each point $p_i$ is projected via spherical projection
to a range-view grid cell $r=\pi(p_i)=(u_i,v_i)$,
and $\mathrm R$ denotes the set of active cells.

Given a pretrained backbone, we perform test-time adaptation (TTA)
by updating lightweight prompt adapters using pseudo-label supervision
on incoming test scans, while keeping the backbone frozen.
Unlike naive TTA~\cite{wang2020tent,wang2022continual},
which updates shared parameters uniformly,
our method restricts adaptation to geometrically supported regions.

As illustrated in Fig.~\ref{fig:structure},
we enforce a geometry-constrained update principle:
(i) spatial adaptation is driven only by supported cells (Sec.~\ref{sec:geo_observability}),
(ii) updates are confined to a lightweight prompt subspace
while the backbone remains frozen (Sec.~\ref{sec:dynamic_prompting}), and
(iii) temporal consistency regularizes adaptation across frames (Sec.~\ref{sec:self_refine}).

\subsection{Beam-Termination and Neighborhood-Supported Observability}
\label{sec:geo_observability}

As illustrated in Fig.~\ref{fig:structure}(A),
we define an observability operator over active cells
to quantify per-cell geometric support.
Under partial LiDAR sensing, returns are sparse,
range-dependent, and affected by occlusion,
resulting in highly non-uniform geometric evidence
across spatial locations~\cite{milioto2019rangenet++,xu2025frnet}.
We therefore construct observability as a quantitative measure
of sensing support for each active cell.

\begin{figure}[t]
    \centering
    \includegraphics[width=1\linewidth]{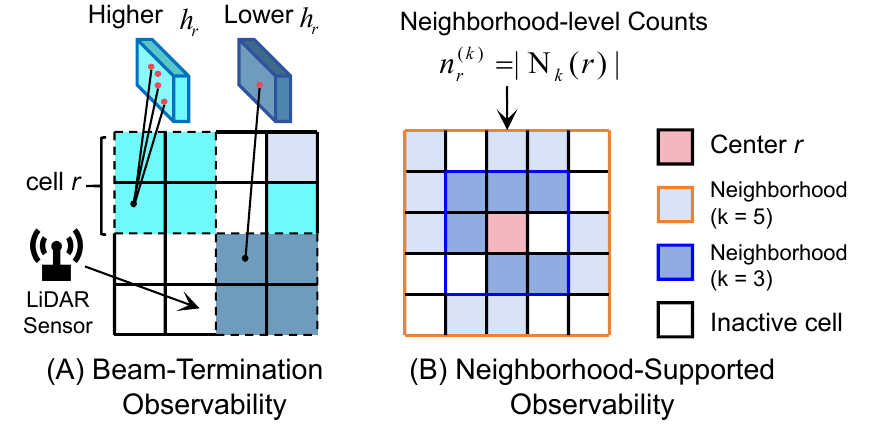}
\caption{(A) Beam-termination and (B) neighborhood-supported observability.
Beam-level return count $h_r$ provides surface evidence,
while neighborhood support counts active cells in
multi-scale Chebyshev neighborhoods centered at $r$.}
    \label{fig:obs_cube}
\end{figure}

\noindent\textbf{Beam-Termination Observability.}
Beam-level observability evaluates surface support
from depth-consistent terminations along sensing rays,
as illustrated in Fig.~\ref{fig:obs_cube}(A).
Cells receiving multiple depth-consistent returns indicate stable surfaces,
whereas mixed-depth returns at occlusion boundaries are unreliable.
Similar to occupancy mapping~\cite{hornung2013octomap},
we interpret depth-consistent termination events as surface evidence.

Since FRNet~\cite{xu2025frnet} retains all points within each frustum region,
per-cell return statistics can be directly computed.
For each $r \in \mathrm R$, let $d_r$ denote the representative depth of cell $r$.
We compute the depth-consistent return count as
\begin{equation}
h_r = \left|\left\{ p_i \in P \mid \pi(p_i)=r,\; |d_i - d_r| < \tau_d \right\}\right|.
\end{equation}

Under fixed angular resolution scanning, LiDAR point density
decreases with range~\cite{benedek2021positioning}.
Since $h_r$ proxies local surface evidence,
identical termination counts imply weaker geometric support
at longer ranges.
To correct this range bias, we define a range-aware support surrogate:
\begin{equation}
s_r = \log(h_r + \epsilon) - \lambda \log(d_r + \epsilon),
\end{equation}
where $\epsilon>0$ ensures numerical stability and $\lambda>0$
controls the strength of range penalization.
This formulation increases with return count while softly
penalizing long-range sparsity.

Since the scale of $s_r$ varies across frames,
we apply median–IQR normalization~\cite{maronna2019robust}
over $\mathrm R$:
\begin{equation}
\psi(x_r)=
\frac{x_r - \mathrm{med}_{q \in \mathrm{R}}(x_q)}
{\mathrm{IQR}_{q \in \mathrm{R}}(x_q) + \epsilon},
\end{equation}
where $\mathrm{med}(\cdot)$ and $\mathrm{IQR}(\cdot)$
denote the per-frame median and interquartile range
for robust scaling.
Finally, we bound the normalized support to $(0,1)$:
\begin{equation}
o_r^{\text{beam}}
=
\sigma\!\big(\psi(s_r)\big),
\end{equation}
where $\sigma(\cdot)$ is the sigmoid function.
This yields the beam-level geometric support $o_r^{\text{beam}}$.

\noindent\textbf{Neighborhood-Supported Observability.}
Beam-termination support alone may overestimate isolated cells.
Semantic segmentation relies on locally coherent geometric structures;
thus reliable regions should exhibit neighborhood continuity.
As illustrated in Fig.~\ref{fig:obs_cube}(B),
for each active cell $r=(u_r,v_r)$ and scale $k\in\mathrm S$,
we define a Chebyshev neighborhood:
\begin{equation}
\mathrm{N}_k(r)=
\{q \in \mathrm{R}\mid \|(u_q,v_q)-(u_r,v_r)\|_\infty
\le \lfloor k/2 \rfloor\}.
\end{equation}
Let $n_r^{(k)}=|\mathrm{N}_k(r)|$ denote
the number of active neighbors.

We aggregate multi-scale neighborhood support as
\begin{equation}
o_r^{\text{neigh}}=
\frac{1}{|\mathrm S|}
\sum_{k\in\mathrm S}
\sigma\!\big(\log(1+n_r^{(k)})\big),
\end{equation}
where $\mathrm S$ contains neighborhood sizes
(e.g., $\{3,5\}$).
Multi-scale aggregation balances sensitivity to thin structures
and robustness to local noise.

\noindent\textbf{Geometry-Consistent Observability.}
We combine beam-termination support and neighborhood support
to obtain a unified sensing-validity score:
\begin{equation}
o_r = o_r^{\text{beam}}\cdot o_r^{\text{neigh}}.
\end{equation}
The multiplicative formulation enforces conjunction:
a cell is geometrically supported only when both
termination evidence and neighborhood continuity are strong.
If either component is weak, the combined support is suppressed.

\noindent\textbf{Efficient Range Statistics.}
The statistics $(h_r,d_r)$ are obtained directly from
the projected range-view representation.
For each active cell, $d_r$ is the stored depth value,
and $h_r$ counts the depth-consistent projected returns.
Computation requires a single linear pass over input points,
yielding $\mathrm O(N)$ complexity per frame.

\subsection{Observability-Constrained Dynamic Prompt Tuning}
\label{sec:dynamic_prompting}

To ensure stable test-time adaptation under sparse and uneven
geometric support, we confine updates to lightweight prompt
adapters while keeping the backbone frozen.

Specifically, prompt adapters are inserted into each residual
stage (Fig.~\ref{fig:structure}(B)),
and the observability score $o_r$ modulates the update strength
across spatial regions.
Unlike conventional static prompt tuning~\cite{yang2022prompt,jia2022visual},
we generate input-dependent prompts conditioned on current features.

\noindent\textbf{Prompt Adapter.}
We insert a lightweight prompt adapter into each residual stage
of the frozen backbone.
Let $F_{i-1}$ denote the feature map before the $i$-th stage.
The prompt is dynamically generated as
\begin{equation}
C_i =
\phi_i\!\left(
\mathrm{LN}(F_{i-1})
\right),
\end{equation}
where $\phi_i(\cdot)$ is a $1\times1$ conv adapter.
This produces an input-conditioned residual calibration
while keeping backbone parameters fixed.

\noindent\textbf{Observability-Modulated Attention.}
The prompt modulates backbone features via a lightweight
per-location cross-attention without spatial interaction.

For each position $r$, we compute a prompt-induced residual
\begin{equation}
\Delta_i^{\text{prompt}}(r)
=
\mathrm{Softmax}\!\left(
\frac{Q_r K_r^\top}{\sqrt d}
\right)
V_r,
\end{equation}
where $Q_r$ is derived from $\mathrm{LN}(F_{i-1}(r))$ and
$K_r, V_r \in \mathbb{R}^{p\times d}$ are generated from
$\mathrm{LN}(C_i(r))$ with a small prompt size $p$.

The final prompt-induced modulation is obtained by
observability gating:
\begin{equation}
F_i(r)
=
F_{i-1}(r)
+
o_r \cdot \Delta_i^{\text{prompt}}(r).
\end{equation}
where the observability score $o_r$ gates the residual update.
Regions lacking geometrically supported evidence thus induce proportionally
weaker adaptation.

\subsection{Temporal Observability Consistency}
\label{sec:self_refine}

To ensure temporally stable adaptation, we maintain a prototype
memory that accumulates only geometrically supported evidence
filtered by $o_r$ (Fig.~\ref{fig:structure}(C)).

Even with spatial gating, predictions for the same object may
flip across frames due to intermittent visibility,
a temporal inconsistency commonly observed in sequential
recognition and video segmentation~\cite{laine2016temporal,li2023tcovis}.
Such flicker introduces unstable supervision signals
for prompt updates.

\noindent\textbf{Prototype-Based Temporal Refinement.}
Let $F_L$ denote the final backbone feature map.
For each cell $r$, we extract its feature
$\mathbf f_r = (F_L)_r$ and obtain the prediction $\mathbf p_r = \mathrm{softmax}(h(\mathbf f_r)),$
where $h(\cdot)$ is the frozen classifier and
$\hat y_r=\arg\max_c \mathbf p_r[c]$.

We maintain class-wise prototypes $\{\mathbf m_c\}_{c=1}^C$
updated via EMA~\cite{tarvainen2017mean,morales2024exponential}
to suppress stochastic noise:
\begin{equation}
\begin{aligned}
\mathbf m_c \leftarrow {}&(1-\beta)\mathbf m_c \\
&+\beta\,
\frac{\sum_{r:\hat y_r=c}
\mathds{1}[\max \mathbf p_r \ge \tau_p]\,
o_r\,(\max \mathbf p_r)\,\mathbf f_r}
{\sum_{r:\hat y_r=c}
\mathds{1}[\max \mathbf p_r \ge \tau_p]\,
o_r\,(\max \mathbf p_r)+\epsilon}.
\end{aligned}
\end{equation}
where $\beta$ controls smoothing and $\tau_p$ is a confidence threshold.
Prototypes are $\ell_2$-normalized after each update:
$\mathbf m_c \leftarrow
\mathbf m_c / (\|\mathbf m_c\| + \epsilon).$

Using the observability-filtered prototypes,
we construct a prototype-aligned target distribution $\mathbf q_r$
via cosine similarity:

\begin{equation}
\begin{aligned}
\mathbf q_r
&=
\mathrm{softmax}\Big(\{\langle \bar{\mathbf f}_r, \bar{\mathbf m}_c \rangle\}_{c=1}^C\Big),\\
\bar{\mathbf f}_r
&=\mathbf f_r/\|\mathbf f_r\|, \;
\bar{\mathbf m}_c=\mathbf m_c/\|\mathbf m_c\|.
\end{aligned}
\end{equation}
The resulting $\mathbf q_r$ defines a temporally consistent alignment target,
which reduces prediction variance induced by fluctuating geometric support.

\noindent\textbf{Prototype-Aligned Prompt Update.}
The prototype-aligned target $\mathbf q_r$ provides
temporally stable supervision for prompt adaptation.
To prevent overconfident or aligned predictions
from dominating updates, we define a reliability weight:
\begin{equation}
w_r =
\max_c \mathbf q_r[c]\,
\Big(1-\mathbf p_r\big[\arg\max_c \mathbf q_r[c]\big]\Big).
\end{equation}

The prompt adapters are updated by minimizing

\begin{equation}
\mathcal{L_\text{TTA}}
=
\sum_{r\in\mathrm R}
o_r\, w_r\,
D_{\mathrm{KL}}\!\big(\mathbf q_r \,\|\, \mathbf p_r\big),
\end{equation}
where $o_r$ gates supervision by geometric observability.
Gradients are propagated exclusively to the prompt adapters,
while the backbone and classifier remain frozen.
Thus, test-time adaptation refines the global prompt
within a low-capacity subspace,
guided by spatial observability
and regularized by temporal prototype consistency.

%% file: sec/4_exps.tex
\section{Experiments}

\begin{table*}
\centering
\caption{SemanticKITTI test set results (mIoU, \%) for LiDAR semantic segmentation across all 19 classes, comparing 2D projection-based and 3D voxel-based methods. \textbf{Bold}: best; \uline{Underlined}: second-best.}
\label{tab:semantickitti}
\setlength{\tabcolsep}{4pt}
\renewcommand{\arraystretch}{1.1}
\resizebox{\linewidth}{!}{
\begin{tabular}{l|c|ccccccccccccccccccl} 
\hline
Method 
& mIoU
& \rotatebox{70}{car}
& \rotatebox{70}{bicycle}
& \rotatebox{70}{motorcycle}
& \rotatebox{70}{truck}
& \rotatebox{70}{other-veh.}
& \rotatebox{70}{person}
& \rotatebox{70}{bicyclist}
& \rotatebox{70}{motorcyclist}
& \rotatebox{70}{road}
& \rotatebox{70}{parking}
& \rotatebox{70}{sidewalk}
& \rotatebox{70}{other-ground}
& \rotatebox{70}{building}
& \rotatebox{70}{fence}
& \rotatebox{70}{vegetation}
& \rotatebox{70}{trunk}
& \rotatebox{70}{terrain}
& \rotatebox{70}{pole}
& \rotatebox{70}{traffic-sign} \\
\hline
\multicolumn{21}{c}{\textbf{3D Voxel Based}}                                                                      \\ 
\hline
Cylinder3D~\cite{zhu2021cylindrical}             & 67.8          & 97.1          & 67.6          & 64.0          & 59.0          & 58.6          & 73.9          & 67.9          & 36.0          & 91.4          & 65.1          & 75.5          & 32.3          & 91.0          & 66.5          & 85.4          & 71.8          & 68.5          & 62.6          & 65.6           \\
SphereFormer~\cite{lai2023spherical}     & 74.8 & 97.5 & 70.1 & 70.5 & 59.6 & 67.7 & 79.0 & 80.4 & 75.3 & 91.8 & 69.7 & 78.2 & 41.3 & 93.8 & 72.8 & 86.7 & 75.1 & 72.4 & 66.8 & 72.9 \\
\hline
\multicolumn{21}{c}{\textbf{2D Projection Based}}                                                                                                                                                                                                                                                                                                       \\ 
\hline
RangeNet++~\cite{milioto2019rangenet++}             & 52.2          & 91.4          & 25.7          & 34.4          & 25.7          & 23.0          & 38.3          & 38.8          & 4.8           & 91.8          & 65.0          & 75.2          & 27.8          & 87.4          & 58.6          & 80.5          & 55.1          & 64.6          & 47.9          & 55.9           \\
PolarNet~\cite{zhang2020polarnet}               & 54.3          & 93.8          & 40.3          & 30.1          & 22.9          & 28.5          & 43.2          & 40.2          & 5.6           & 90.8          & 61.7          & 74.4          & 21.7          & 90.0          & 61.3          & 84.0          & 65.5          & 67.8          & 51.8          & 57.5           \\
RangeViT~\cite{ando2023rangevit}               & 64.0          & 95.4          & 55.8          & 43.5          & 29.8 & 42.1          & 63.9          & 58.2          & 38.1          & \textbf{93.1}          & 70.2          & 80.0          & 32.5& 92.0 & 69.0 & 85.3          & 70.6          & 71.2          & 60.8          & 64.7           \\
SFCNet~\cite{zheng2024spherical} & 65.0 & 95.1          & 64.2 & 63.2 & 23.5          & 45.6 & 78.3 & 73.1 & 26.4          & 87.9          & 65.6          & 71.9          & 29.1          & 91.1          & 64.5          & 83.7          & 72.6 & 69.6          & 62.6 & 67.2           \\
FRNet~\cite{xu2025frnet} & 73.3 & 97.3 &67.9 &74.6 &59.4 &\textbf{66.3} &78.1 &\textbf{79.2} &57.3 &92.1 &73.0 &78.1 &41.8 &\textbf{92.7} &71.0 &86.7                &73.2 &72.5 &64.7 &67.3           \\
\textbf{Ours} & \textbf{74.3} & \textbf{97.5} & \textbf{69.1} & \textbf{75.9} & \textbf{61.1} & \uline{65.8} & \textbf{79.4} & \textbf{79.2} & \textbf{58.9} & \uline{93.0} & \textbf{74.8} & \textbf{79.8} & \textbf{42.4} & \uline{92.0} & \textbf{72.4} & \textbf{87.2} & \textbf{74.7} & \textbf{73.2} & \textbf{66.2} & \textbf{68.5}
           \\
\hline
\end{tabular}
}
\end{table*}

\begin{table*}
\centering
\caption{nuScenes test set results (mIoU, \%) for LiDAR semantic segmentation across all 16 classes, comparing 2D projection-based and 3D voxel-based methods. \textbf{Bold}: best; \uline{Underlined}: second-best.}
\label{tab:nuscenes}
\setlength{\tabcolsep}{3pt}
\renewcommand{\arraystretch}{1.1}
\resizebox{\linewidth}{!}{
\begin{tabular}{l|c|c|cccccccccccccccc}
\hline
Method 
& Venue
& mIoU
& \rotatebox{70}{barrier}
& \rotatebox{70}{bicycle}
& \rotatebox{70}{bus}
& \rotatebox{70}{car}
& \rotatebox{70}{const.-veh.}
& \rotatebox{70}{motorcycle}
& \rotatebox{70}{pedestrian}
& \rotatebox{70}{traffic-cone}
& \rotatebox{70}{trailer}
& \rotatebox{70}{truck}
& \rotatebox{70}{driv.-surf.}
& \rotatebox{70}{other-ground}
& \rotatebox{70}{sidewalk}
& \rotatebox{70}{terrain}
& \rotatebox{70}{manmade}
& \rotatebox{70}{vegetation} \\
\hline

\multicolumn{19}{c}{\textbf{3D Voxel Based}} \\
\hline
Cylinder3D~\cite{zhu2021cylindrical}
& CVPR'21
& 76.1 & 76.4 & 40.3 & 91.2 & 93.8 & 51.3 & 78.0 & 78.9 & 64.9 & 62.1 & 84.4 & 96.8 & 71.6 & 76.4 & 75.4 & 90.5 & 87.4 \\

SphereFormer~\cite{lai2023spherical}
& CVPR'23
& 78.4 & 77.7 & 43.8 & 94.5 & 93.1 & 52.4 & 86.9 & 81.2 & 65.4 & 73.4 & 85.3 & 97.0 & 73.4 & 75.4 & 75.0 & 91.0 & 89.2 \\
\hline

\multicolumn{19}{c}{\textbf{2D Projection Based}} \\
\hline
RangeNet++~\cite{milioto2019rangenet++}
& IROS'19
& 65.5 & 66.0 & 21.3 & 77.2 & 80.9 & 30.2 & 66.8 & 69.6 & 52.1 & 54.2 & 72.3 & 94.1 & 66.6 & 63.5 & 70.1 & 83.1 & 79.8 \\

PolarNet~\cite{zhang2020polarnet}
& CVPR'20
& 71.0 & 74.7 & 28.2 & 85.3 & 90.9 & 35.1 & 77.5 & 71.3 & 58.8 & 57.4 & 76.1 & 96.5 & 71.1 & 74.7 & 74.0 & 87.3 & 85.7 \\

RangeViT~\cite{ando2023rangevit}
& CVPR'23
& 75.2 & 75.5 & 40.7 & 88.3 & 90.1 & 49.3 & 79.3 & 77.2 & 66.3 & 65.2 & 80.0 & 96.4 & 71.4 & 73.8 & 73.8 & 89.9 & 87.2 \\

SFCNet~\cite{zheng2024spherical}
& NeurIPS'24
& 75.9 & 76.7 & 40.4 & 89.5 & 91.3 & 46.7 & 82.0 & 78.1 & 65.8 & 69.4 & 80.6 & 96.6 & 71.6 & 74.5 & 74.9 & 89.0 & 87.5 \\

FRNet~\cite{xu2025frnet}
& TIP'25
& 79.0 & 78.5 & 43.9 & \textbf{95.4} & 93.2 & 56.3 & 85.8 & 79.0 & 68.5 & 72.8 & \textbf{86.5} & \textbf{97.1} & 75.9 & 77.0 & 76.4 & 89.7 & 88.0 \\

\textbf{Ours}
& IROS'26
& \textbf{80.2} & \textbf{79.7} & \textbf{45.6} & \uline{95.0} & \textbf{94.7} & \textbf{57.6} & \textbf{87.1} & \textbf{80.7} & \textbf{69.4} & \textbf{74.0} & \uline{86.1} & \uline{96.8} & \textbf{77.6} & \textbf{78.9} & \textbf{77.8} & \textbf{91.3} & \textbf{90.1} \\
\hline
\end{tabular}
}
\end{table*}

\subsection{Experimental Setup}
\noindent\textbf{Datasets.}
We evaluate on SemanticKITTI~\cite{behley2019semantickitti} and nuScenes~\cite{caesar2020nuscenes} following their official splits.
SemanticKITTI contains 43,551 scans annotated with 19 classes, using sequences 00--07 and 09--10 for training, 08 for validation, and 11--21 for testing.
nuScenes includes 34,149 scans with point-wise annotations over 16 classes.

\noindent\textbf{Evaluation Metrics.}
We evaluate LiDAR semantic segmentation performance using mean Intersection-over-Union (mIoU)
over all semantic classes, following the standard protocol adopted in prior
segmentation works~\cite{ando2023rangevit,zheng2024spherical}.

\noindent\textbf{Implementation Details.}
We initialize the backbone with official FRNet~\cite{xu2025frnet} pretrained weights.
The prompt adapters are trained on the official training split for 10 epochs using Adam with a learning rate of $1\times10^{-4}$, while keeping the backbone frozen.
After supervised prompt training, we perform online test-time adaptation (TTA).
The backbone remains frozen, and only the lightweight residual adapters are updated per frame using a learning rate of $5\times10^{-4}$.
We use $\lambda=0.5$, EMA decay $\beta=0.99$, and neighborhood sizes $\mathrm{S}=\{3,5\}$.
All experiments are conducted on a single NVIDIA A100 (40GB).

\vspace{10pt}

\subsection{Comparison with State-of-the-Art Methods}
\noindent\textbf{3D vs.\ 2D Baselines.} 
We compare our method with both 3D voxel-based approaches
(Cylinder3D~\cite{zhu2021cylindrical}, SphereFormer~\cite{lai2023spherical})
and 2D projection-based methods
(RangeNet++~\cite{milioto2019rangenet++}, PolarNet~\cite{zhang2020polarnet},
RangeViT~\cite{ando2023rangevit}, SFCNet~\cite{zheng2024spherical}, 
FRNet~\cite{xu2025frnet}).

Our method builds upon the 2D projection-based backbone FRNet for test-time prompt tuning.
This design avoids the dynamic sparse topology of 3D voxel-based networks, which results in input-dependent topology and higher computational overhead during online adaptation.
The fixed spatial topology and deterministic geometric indexing in 2D projections facilitate stable optimization and efficient updates.

\noindent\textbf{Quantitative Results on SemanticKITTI.}
Table~\ref{tab:semantickitti} reports results on the SemanticKITTI test set.
Our method achieves the highest mIoU among 2D projection-based approaches and significantly narrows the performance gap to 3D voxel-based backbones, even surpassing them on several classes.
In particular, we observe clear improvements on geometrically sparse classes such as \emph{person}, \emph{bicyclist}, and \emph{motorcyclist}, as observability-guided tuning suppresses unreliable gradients and stabilizes features on thin and distant structures.

\begin{figure*}
    \centering
    \includegraphics[width=1\linewidth,height=0.6\textheight]{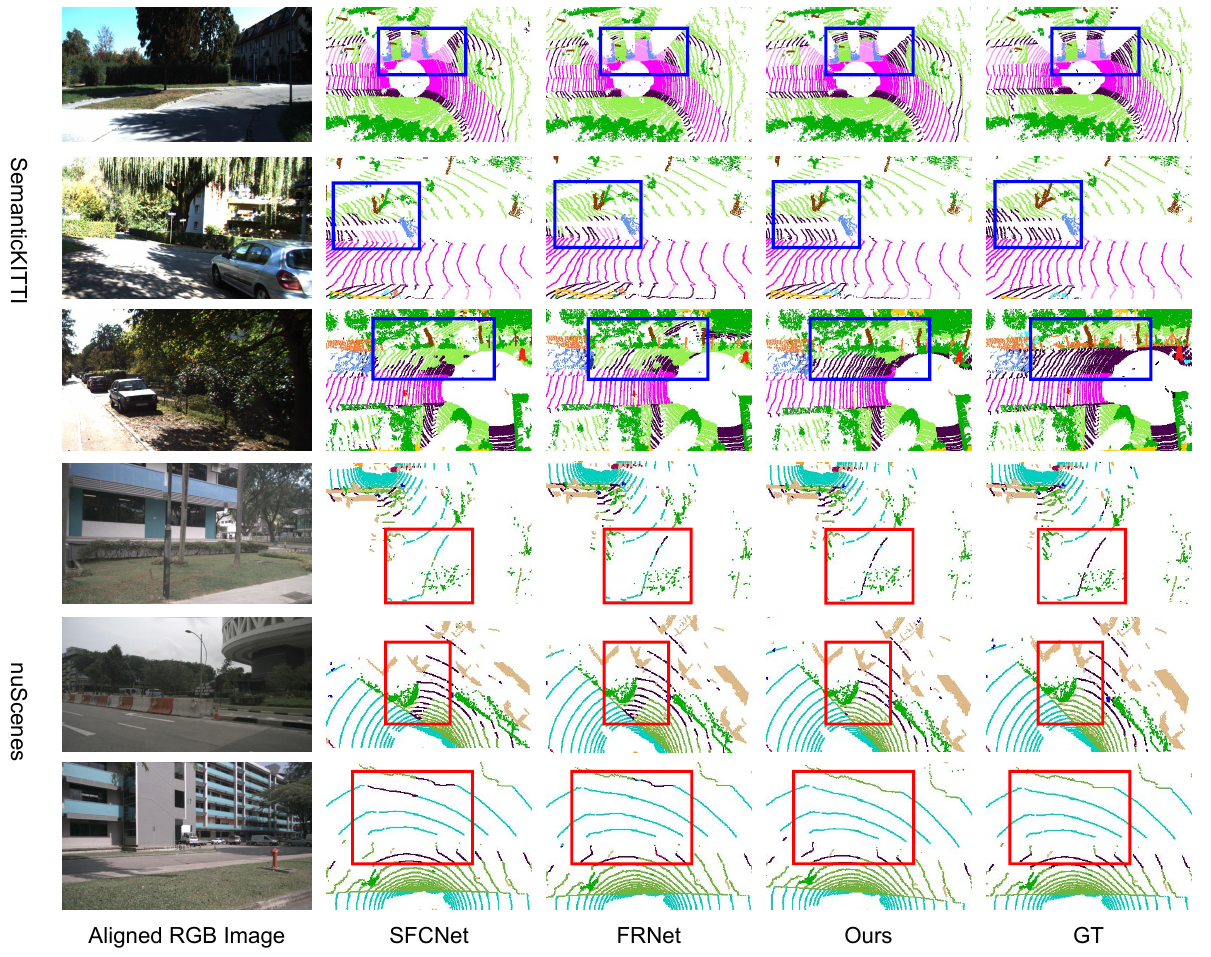}
\caption{
Qualitative comparison on SemanticKITTI (Rows 1--3) and nuScenes (Rows 4--6).
From left to right: RGB image, SFCNet~\cite{zheng2024spherical}, FRNet~\cite{xu2025frnet}, ours, and ground truth.
Our method produces cleaner planar regions, less fragmentation on thin structures, and more coherent boundaries. \bluetext{Blue} and \textcolor{red}{Red} boxes highlight representative differences.
}
    \label{fig:compare}
\end{figure*}

\noindent\textbf{Quantitative Results on nuScenes.}
Table~\ref{tab:nuscenes} reports results on the nuScenes test set.
Our method achieves the highest mIoU among 2D projection-based approaches and surpasses 3D voxel-based backbones in overall performance.
Improvements are consistently observed across both large-scale classes (e.g., \emph{car}, \emph{truck}) and small or irregular objects (e.g., \emph{traffic-cone}), indicating stable performance across diverse object scales and scene layouts.

\noindent\textbf{Cross-Dataset TTA Results.}
Table~\ref{tab:cross} reports cross-dataset adaptation results on SemanticKITTI $\rightarrow$ nuScenes and nuScenes $\rightarrow$ SemanticKITTI, reflecting real-world deployment scenarios where models must adapt to different sensing conditions and annotation distributions without supervision.
We compare our approach with representative TTA methods including Tent~\cite{wang2020tent}, GIPSO~\cite{saltori2022gipso}, and HGL~\cite{zou2024hgl}.

Our method achieves the best mIoU in both transfer directions, improving over the source-only model by \textbf{+3.09\%} on KITTI$\rightarrow$nuScenes and \textbf{+9.32\%} on nuScenes$\rightarrow$KITTI.
Despite updating only \textbf{0.72\%} of the parameters, our approach surpasses methods that adapt the full backbone (\textbf{100\%}) while achieving substantially lower per-frame adaptation time (\textbf{0.03s}), demonstrating efficient geometry-constrained cross-domain adaptation.

\noindent\textbf{Qualitative Results on SemanticKITTI and nuScenes.}
Fig.~\ref{fig:compare} shows qualitative comparisons on SemanticKITTI (Rows 1--3)
and nuScenes (Rows 4--6) against SFCNet~\cite{zheng2024spherical}
and FRNet~\cite{xu2025frnet}. Blue and red boxes highlight key differences
in challenging regions with sparse observations.
\begin{enumerate}[label=\arabic*),leftmargin=*,itemsep=2pt]
\item \textbf{SemanticKITTI (Rows 1--3).}
Our method produces cleaner planar regions (e.g., \emph{road}, \emph{parking}, \emph{sidewalk})
with sharper class boundaries and more consistent surface predictions.
Thin structures such as \emph{fence} are better preserved,
while competing methods exhibit label bleeding and fragmented segments.

\item \textbf{nuScenes (Rows 4--6).}
Our approach reduces isolated errors on \emph{barrier},
improves structural continuity on large regions (e.g., \emph{road}, \emph{sidewalk}),
and yields fewer scattered misclassifications in vegetation and terrain,
resulting in more coherent scene layouts and more stable predictions.
\end{enumerate}

\begin{table}[H]
\centering
\caption{Cross-dataset TTA results in mIoU (\%).
Params. denotes the percentage of trainable parameters during test-time adaptation.
Time (s) reports the average per-frame adaptation time measured on SemanticKITTI.}
\setlength{\tabcolsep}{3pt}
\renewcommand{\arraystretch}{1.1}
\resizebox{\linewidth}{!}{
\begin{tabular}{lcccc}
\toprule
Setting & KITTI$\rightarrow$nuScenes  &nuScenes $\rightarrow$KITTI & Params. & Time(s) \\
\hline
Source & 32.03 & 37.40 & - & - \\
Tent~\cite{wang2020tent} & 30.62 (-1.41) & 38.06 (+0.66) & 100\% & 0.37 \\
GIPSO~\cite{saltori2022gipso} & 33.35 (+1.05) & 40.28 (+2.88) & 100\% & 3.21 \\
HGL~\cite{zou2024hgl} & 34.49 (+2.46) & 45.06 (+7.66) & 100\% & 0.55 \\
\textbf{Ours} & 35.12 \textbf{(+3.09)} & 46.72 \textbf{(+9.32)} & 0.72\% & 0.03 \\
\bottomrule
\end{tabular}
}
\label{tab:cross}
\end{table}

\subsection{Ablation Studies}

\noindent\textbf{Plug-and-Play Prompt Ablation across Backbones.}
Table~\ref{tab:backbone} evaluates three representative backbones:
RangeViT~\cite{ando2023rangevit},
SFCNet~\cite{zheng2024spherical}, and
FRNet~\cite{xu2025frnet}.
Our module can be inserted in a plug-and-play manner
without modifying the original architecture.

For fair comparison, we implement VPT~\cite{jia2022visual}
by freezing backbone and injecting $M{=}5$ prompt tokens
($C{=}128$) via feature concatenation followed by
a linear projection:
$\tilde f = W [f ; p],$
with identity-initialized $W$.

Across all backbones, our method achieves larger mIoU gains
than vanilla VPT while requiring substantially fewer parameters,
with only 0.06–0.09M additional parameters.

It improves performance by +2.1 / +2.4 mIoU on RangeViT,
+0.9 / +1.5 on SFCNet, and +1.0 / +1.2 on FRNet
under SemanticKITTI / nuScenes, 
demonstrating the effectiveness of explicitly modeling
LiDAR observability and the generality of our backbone-agnostic design.

\begin{table}
\centering
\caption{Plug-and-play comparison of vanilla VPT~\cite{jia2022visual} and our observability-aware prompting
across three backbones on SemanticKITTI and nuScenes.}
\label{tab:backbone}
\resizebox{\linewidth}{!}{
\begin{tabular}{clccc} 
\hline
                                         & Method    & SemanticKITTI & nuScenes      & Params. (M)  \\ 
\hline
\multirow{3}{*}{(A)}                     & RangeViT~\cite{ando2023rangevit} & 55.8          & 75.2          & 23.7         \\
                                         & + VPT~\cite{jia2022visual}     & 56.4          & 76.1          & +0.71       \\
                                         & + Ours    & \textbf{57.9} & \textbf{77.6} & +0.09       \\ 
\hline
\multirow{3}{*}{(B)}                     & SFCNet~\cite{zheng2024spherical}   & 64.2          & 75.9          & 11.4         \\
                                         & + VPT~\cite{jia2022visual}     & \textbf{65.5} & 76.8          & +0.49       \\
                                         & + Ours    & \uline{65.1}  & \textbf{77.4} & +0.06       \\ 
\hline
\multicolumn{1}{l}{\multirow{3}{*}{(C)}} & FRNet~\cite{xu2025frnet}    & 73.3          & 79.0          & 10.0         \\
\multicolumn{1}{l}{}                     & + VPT~\cite{jia2022visual}     & 73.9          & 79.6          & +0.56       \\
\multicolumn{1}{l}{}                     & + Ours    & \textbf{74.3} & \textbf{80.2} & +0.07       \\
\hline
\end{tabular}
}
\end{table}

\noindent\textbf{Impact of Individual Components.}
Table~\ref{tab:ablate} decomposes our framework into
prompt adaptation, geometric observability, and temporal prototype consistency.
Prompt adaptation alone (A→B) already improves performance,
confirming that parameter-efficient feature recalibration
mitigates distribution shift.
However, without observability constraints, updates are still driven
by all spatial locations.
Introducing geometric observability (B→C) further improves mIoU,
verifying that restricting gradient updates to geometrically supported regions
stabilizes adaptation under partial LiDAR sensing.
Finally, adding temporal prototype consistency (C→E)
achieves the best performance (74.3 / 80.2),
demonstrating that spatially reliable updates and temporally stable targets
act synergistically to enhance test-time robustness.

\begin{table}
\centering
\caption{
Ablation of geometric observability, prompt adaptation,
and temporal prototype consistency on SemanticKITTI and nuScenes.
}
\setlength{\tabcolsep}{3pt}
\renewcommand{\arraystretch}{1.1}
\label{tab:ablate}
\resizebox{\linewidth}{!}{
\begin{tabular}{l c c c c c}
\hline
 & Observability & Prompt & Temporal Proto. & SemanticKITTI & nuScenes \\
\hline
(A) &  &  &  & 73.3 & 79.0 \\
(B) &  & \checkmark &  & 73.8 & 79.6 \\
(C) & \checkmark & \checkmark &  & 74.1 & 80.0 \\
(D) &  & \checkmark & \checkmark & 73.9 & 79.8  \\
(E) & \checkmark & \checkmark & \checkmark & 74.3 & 80.2 \\
\hline
\end{tabular}
}
\end{table}

\begin{figure}
    \centering
    \includegraphics[width=1\linewidth]{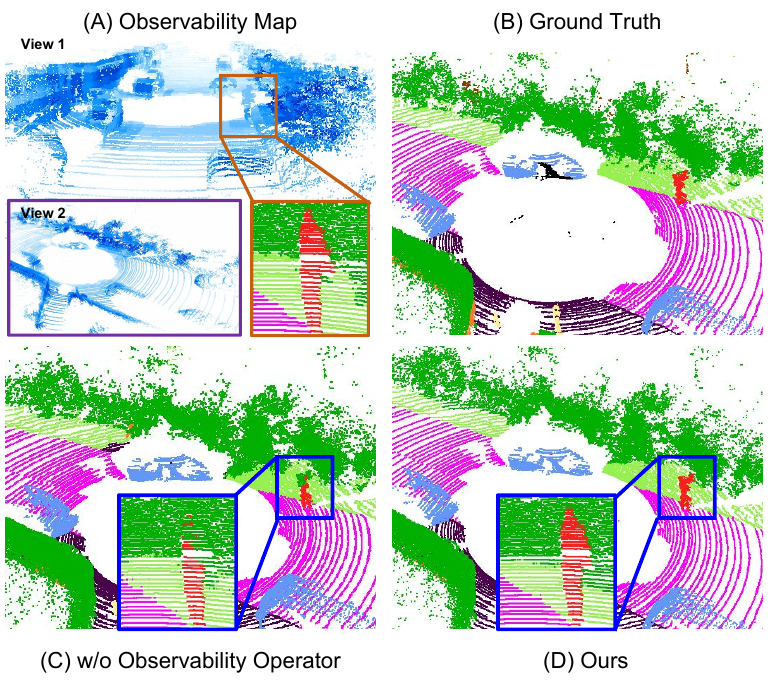}
\caption{
Visualization of explicit observability modeling.
(A) Observability map from two viewpoints, encoding per-cell geometric support.
(B) Ground truth.
(C) Without observability modeling, weakly supported regions yield fragmented predictions.
(D) Our method, driven by geometrically supported cells, recovers a coherent pedestrian structure.
}
    \label{fig:ablate}
\end{figure}

\noindent\textbf{Effect of Explicit Observability Modeling.}
Fig.~\ref{fig:ablate}(A) visualizes the observability map,
which encodes the per-cell geometric support computed from sensing evidence.
Without observability modeling, gradients from weakly supported cells
can influence shared parameters, leading to fragmented or distorted predictions
(e.g., Fig.~\ref{fig:ablate}(C) the incomplete pedestrian structure in w/o observability).

By modeling observability and down-weighting unreliable cells,
adaptation is primarily driven by geometrically supported regions.
Consequently, the pedestrian structure is recovered more coherently
and better aligned with GT.

The quantitative improvement (B→C in Table~\ref{tab:ablate})
further confirms that observability modeling effectively
stabilizes adaptation under partial LiDAR sensing.

\noindent\textbf{Computational Efficiency.}
As shown in Table~\ref{tab:cross} and Table~\ref{tab:ablate}, 
our method updates only a negligible fraction of model parameters
(\textbf{0.5\%–0.7\% }of backbone parameters).
On an NVIDIA A100 GPU, ours requires 0.03 s per frame,
compared to 0.55 s for full-parameter adaptation,
corresponding to an \textbf{18× reduction} in per-frame adaptation time.

% Despite this reduction, our method achieves higher mIoU
% across all evaluated settings,
% indicating that geometry-constrained prompt tuning
% reduces adaptation cost without sacrificing performance.

%% file: sec/con.tex
\vspace{10pt}
\section{Conclusion}

We present a geometry-constrained test-time prompt tuning framework for LiDAR semantic segmentation. Motivated by spatially heterogeneous pseudo-label reliability, we show that uniform updates under globally shared parameters can destabilize adaptation. To address this issue, we introduce a geometry-aware observability mechanism that identifies geometrically supported evidence, confine adaptation to a lightweight prompt subspace, and incorporate temporally smoothed prototype alignment to suppress flicker. 
Together, these components ensure that adaptation is driven only by the resulting geometrically supported evidence, enabling stable and efficient online adaptation under evolving sensing conditions without additional annotations.

\vspace{10pt}
\noindent\textbf{Limitation and Future Work.} While this work focuses on LiDAR semantic segmentation, we believe the proposed observability-guided adaptation principle suggests a broader way of reasoning about test-time adaptation: adaptation stability should be governed by what is physically observable, rather than treating all pseudo-labels as equally reliable. Future work will investigate extending this principle to multi-modal perception and other dense prediction tasks.